\documentclass[runningheads]{llncs}
\usepackage{graphicx}

\usepackage{amsmath,amssymb} 
\usepackage{color}
\usepackage{tikz}
\usepackage{scrextend}
\usepackage[normalem]{ulem}
\usepackage{caption}
\usepackage{subcaption}
\usepackage{multirow}
\usepackage{array}
\usepackage{makecell} 
\usetikzlibrary{calc}

\usepackage{hyperref}

\usepackage{etoolbox}
\newcommand{\repthanks}[1]{\textsuperscript{\ref*{#1}}}
\makeatletter
\patchcmd{\maketitle}
  {\def\thanks}
  {\let\repthanks\repthanksunskip\def\thanks}
  {}{}
\patchcmd{\@maketitle}
  {\def\thanks}
  {\let\repthanks\@gobble\def\thanks}
  {}{}
\newcommand\repthanksunskip[1]{\unskip{}}
\makeatother

\begin{document}
\title{The Eyecandies Dataset for Unsupervised Multimodal Anomaly Detection and Localization}
\titlerunning{The Eyecandies Dataset}
%
\author{Luca Bonfiglioli\thanks{Joint first authorship.\protect\label{X}}\orcidID{0000-0002-7323-0662} \and
Marco Toschi\repthanks{X}\orcidID{0000-0002-6573-2857} \and
Davide Silvestri\orcidID{0000-0003-0727-7785} \and
Nicola Fioraio\orcidID{0000-0001-9969-0555} \and
Daniele {De Gregorio}\orcidID{0000-0001-8203-9176}}
\authorrunning{Bonfiglioli et al.}
%
\institute{Eyecan.ai \email{\{luca.bonfiglioli, marco.toschi, davide.silvestri, nicola.fioraio, daniele.degregorio\}@eyecan.ai} \href{https://www.eyecan.ai}{https://www.eyecan.ai}}
\maketitle              
\begin{abstract}
We present Eyecandies, a novel synthetic dataset for unsupervised anomaly detection and localization. Photo-realistic images of procedurally generated candies are rendered in a controlled environment under multiple lightning conditions, also providing depth and normal maps in an industrial conveyor scenario. We make available anomaly-free samples for model training and validation, while anomalous instances with precise ground-truth annotations are provided only in the test set. The dataset comprises ten classes of candies, each showing different challenges, such as complex textures, self-occlusions and specularities. Furthermore, we achieve large intra-class variation by randomly drawing key parameters of a procedural rendering pipeline, which enables the creation of an arbitrary number of instances with photo-realistic appearance. Likewise, anomalies are injected into the rendering graph and pixel-wise annotations are automatically generated, overcoming human-biases and possible inconsistencies.

We believe this dataset may encourage the exploration of original approaches to solve the anomaly detection task, e.g. by combining color, depth and normal maps, as they are not provided by most of the existing datasets. Indeed, in order to demonstrate how exploiting additional information may actually lead to higher detection performance, we show the results obtained by training a deep convolutional autoencoder to reconstruct different combinations of inputs.

\end{abstract}
%
%
%
\section{Introduction}
Recent years have seen an increasing interest in visual unsupervised anomaly detection \cite{ADsurvey}, the task of determining whether an example never seen before presents any aspects that deviate from a defect-free domain, which was learned during training. Similar to one-class classification \cite{moya1996network,KEMMLER20133507,sabokrou2018adversarially}, in unsupervised anomaly detection the model has absolutely no knowledge of the appearance of anomalous structures and must learn to detect them solely by looking at \emph{good} examples. There is a practical reason behind this apparent limitation: being anomalies rare by definition, collecting real-world data with enough examples of each possible deviation from a target domain may prove to be unreasonably expensive. Furthermore, the nature of all possible anomalies might even be unknown, so treating anomaly detection as a supervised classification task may hinder the ability of the model to generalize to new unseen types of defects.

Historically, a common evaluation practice for proposed AD methods was to exploit existing multi-class classification datasets, such as MNIST \cite{MNIST} and CIFAR \cite{CIFAR}, re-labeling a subset of related classes as inliers and the remaining as outliers \cite{Perera_2019_CVPR}. The major drawback of this practice is that clean and anomalous domains are often completely unrelated, whereas in real-world scenarios, such as industrial quality assurance or autonomous driving, anomalies usually appear as subtle changes within a common scene, as for the anomalies presented in \cite{MVTECADLOGIC}. In the recent years this adaptation of classification datasets was discouraged in favor of using new datasets specifically designed for visual anomaly detection and localization, such as \cite{MVTECAD}, which focuses on industrial inspection. However, most of the available datasets provide only color images with ground-truth annotations and very few add 3D information \cite{MVTEC3D}. Moreover, all of them have to face the problem of manual labelling, which can be human-biased and error-prone, especially in the 3D domain.

\begin{figure}
    \centering
    \includegraphics[width=\linewidth]{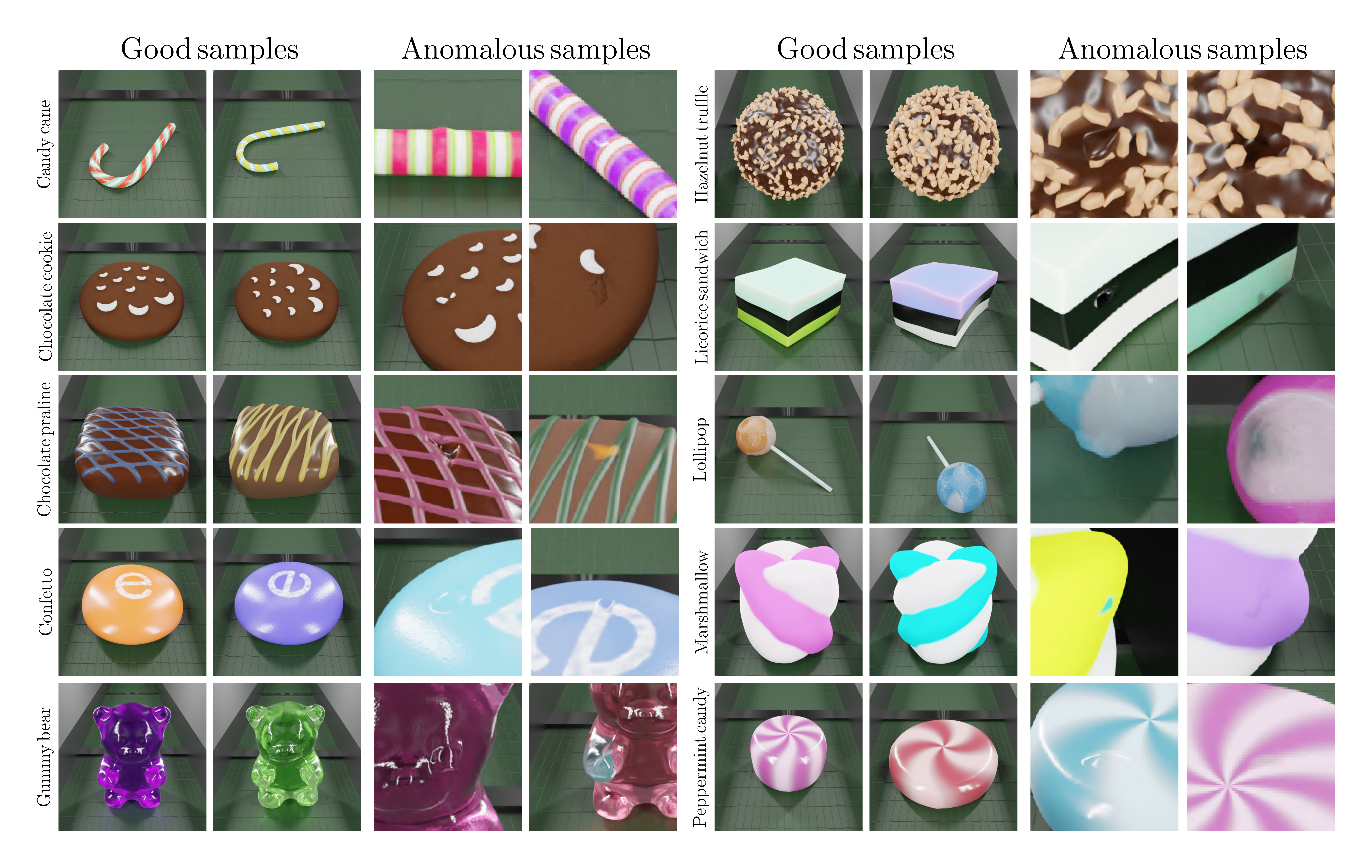}
    \caption{Examples from the Eyecandies dataset. Each row shows good and bad samples from the same object category (best viewed in color).} 
    \label{fig:allexamples}
\end{figure}

The Eyecandies dataset is our main contribution to tackle these issues and provide a new and challenging benchmark for unsupervised anomaly detection, including a total of 90000 photo-realistic shots of procedurally generated synthetic objects, spanning across 10 classes of candies, cookies and sweets (cfr.~Fig.~\ref{fig:allexamples}). Different classes present entirely different shapes, color patterns and materials, while intra-class variance is given by randomly altering parameters of the same model. The Eyecandies dataset comprises defect-free samples for training, as well as anomalous ones used for testing, each of them with automatically generated per-pixel ground-truth labels, thus removing the need for expensive (and often biased) manual annotation procedures. Of each sample, we also provide six renderings with different controlled lighting conditions, together with ground-truth depth and normal maps, encouraging the exploration and comparison of many alternative approaches.

We found that performance of existing methods on synthetic data are in line with the results obtained on \emph{real} data, such as \cite{MVTECAD}, though our dataset appears to be more challenging. Moreover, being the use of 3D data not common in the AD field, we deployed a deep convolutional autoencoder trained to reconstruct different combination of inputs, showing that the inclusion of 3D data results in better anomaly detection and localization performance.

To explore the data and evaluate a method, please go to \href{https://eyecan-ai.github.io/eyecandies}{https://eyecan-ai.github.io/eyecandies}. Please refer to \href{https://github.com/eyecan-ai/eyecandies}{https://github.com/eyecan-ai/eyecandies} for examples and tutorials on how to use the Eyecandies dataset.


\section{Related Work\label{sec:related-works}}

Anomaly detection and localization on images (hereinafter AD) is an ubiquitous theme in many fields, from autonomous driving \cite{Blum2019FishyscapesAB} to visual industrial inspection \cite{MagneticTile,SteelDefectDetection,MVTECAD}. Likewise, the use of synthetic datasets to evaluate the performance of proposed methods has been already explored in many contexts \cite{wang2020tartanair,SyntheticSteel,SCHMEDEMANN20221101,gutierrez2021synthetic}.
However, very few works investigate how synthetic data can be effectively exploited to advance the AD field, which is
indeed the focus of the dataset we are presenting.
In the next sections we will first review the publicly available datasets for AD, then we will briefly analyze the most successful methods proposed to solve the AD task, showing how our work may help the research community.

\subsection{Anomaly Detection Datasets}

Different public AD datasets exists, some designed for industrial inspection of a very specialized type of objects, while other trying to be more generic. An example of the former group is the Magnetic Tile Dataset \cite{MagneticTile}, composed by 952 anomaly-free images and 5 types of anomalies, for a total of 1501 manually annotated images of various resolutions. Despite being a reference in the field, this dataset comprises only a single texture category and it is limited to grayscale images.
Though much larger, \cite{SteelDefectDetection} is another similar dataset presented on Kaggle, focused on a single object class.

The NanoTWICE dataset \cite{NanofibrousMaterials} provides high resolution images (1024x696), although of little interest for deep learning approaches, since it is composed by only 5 anomaly-free images and 40 images with anomalies of different sizes.

In \cite{SynthethicTextures} the authors generate a synthetic dataset of 1000 good images and 150 anomalous images, with ground-truth labels approximated by ellipses. The test set comprises 2000 non defective images and 300 defective ones as 8-bit grayscale with a resolution of 512x512. Though larger than usual, it shows low texture variation and the ground-truth is very coarse. Instead, our synthetic pipeline aims at photo-realistic images with large intra-class variation and pixel-precise ground-truth masks.

MVTec AD \cite{MVTECAD}, which is focused on the industrial inspection scenario, features a total of 5354 real-world images, spanning across 5 texture and 10 object categories.
The test set includes 73 distinct types of anomalies (on average 5 per category) with a total of 1725 images.
Anomalous regions have been manually annotated, though introducing small inconsistencies and unclear resolution of missing object parts. In our work we purposely avoid these undefined situations, while providing pixel-precise annotations in an automated way.

MVTec LOCO AD \cite{MVTECADLOGIC} introduces the concept of ``structural'' and ``logical'' anomalies: the former being local irregularities like scratches or dents, and the latter being violations of underlying logical constraints that require a deeper understanding of the scene. The dataset consists of 3644 images, distributed across 6 categories. Though interesting and challenging, the detection of logical anomalies is out of the scope of this work, where we focus on localized defects only. Moreover, such defects are usually specific for a particular object class, while we aim at automated and consistent defect generation. Finally, being the subject fairly new, there is no clear consensus on how to annotate the images and evaluate the performance of a method.

MVTec 3D-AD \cite{MVTEC3D} has been the first 3D dataset for AD. Authors believe that the use of 3D data is not common in the AD field due to the lack of suitable datasets. They
provide 4147 point clouds, acquired by an industrial 3D sensor, and a complementary RGB image for 10 object categories.
The test set comprises 948 anomalous objects and 41 types of defects, all manually annotated. The objects are captured on a black background, useful for data augmentation, but not very common in real-world scenarios.
Moreover, the use of a 3D device caused the presence of occlusions, reflections and inaccuracies, introducing a source
of noise that may hinder a fair comparison of different AD proposals.
Of course, our synthetic generation does not suffer from such nuisances.

Synthetic generation of defective samples is introduced in \cite{SyntheticSteel} to enhance the performance of an AD classifier.
As in our work, they use Blender \cite{blenderman} to create the new data, though they focus on combining real and synthetic images, while we aim at providing a comprehensive dataset for evaluation and comparison. Also, the authors of \cite{SyntheticSteel} did not release their dataset.

In \cite{SCHMEDEMANN20221101} another non-publicly available dataset is presented. They render 2D images from 3D models in a procedural way, where randomized parameters control defects, illumination, camera poses and texture. Their rendering pipeline is similar to ours, though, as in \cite{SyntheticSteel}, their focus is on generating huge amounts of synthetic anomalous examples to train a model in a supervised fashion that could generalize to real data.

Finally, in \cite{gutierrez2021synthetic} the authors propose to apply synthetic defects on the 3D reconstruction of a target object. The rendering pipeline shares some of our intuitions, such as parametric modeling of defects and rendering.
However, the use of expensive hardware and the need of a physical object hinder general applicability to build up a comprehensive dataset. Moreover the approach followed in this and in the previously cited papers \cite{SyntheticSteel,gutierrez2021synthetic} about synthetic augmentation is different from ours since the model is trained on anomalous data.

\subsection{Methods}

In the last few years, many novel proposals emerged to tackle the AD task. Generally, methods can be categorized as discriminative or generative, where the former often model the distribution of features extracted, e.g., from a pre-trained neural network \cite{PADIM,PATCHCORE,FASTFLOW,STUDENT-TEACHER}, while the latter prefer an end-to-end training \cite{MagneticTile,AutoencoderSSIM,akcay2018ganomaly,FANOGAN}. Due to the lack of diverse 3D dataset for AD, very few proposals are explicitly designed to exploit more than just a single 2D color image \cite{viana20213dbrain,bengs2021three}. Therefore, we give the community a novel dataset to further investigate the use of 3D geometry, normal directions and lighting patterns in the context of AD.

\section{The Eyecandies Dataset\label{sec:the-dataset}}

The Eyecandies dataset comprises ten different categories of candies, chosen to provide a variety of shapes, textures and materials:
\begin{itemize}
    \item Candy Cane
    \item Chocolate Cookie
    \item Chocolate Praline
    \item Confetto
    \item Gummy Bear
    \item Hazelnut Truffle
    \item Licorice Sandwich
    \item Lollipop
    \item Marshmallow
    \item Peppermint Candy
\end{itemize}

Our pipeline generates a large number of unique instances of each object category, all of them differing in some controlled aspects. A subset of samples, labeled as defective, present one or more anomalies on their surface. An automatically annotated ground-truth segmentation mask provides pixel-precise classification labels.

In the next subsection we will generally describe the setup and the data produced by our pipeline, then in Sec.~\ref{sec:defects} we will present the available types of defects. More details on the data generation process are found in Sec.~\ref{sec:data-gen}.

\subsection{General Setup\label{sec:general-setup}}

\begin{figure}
    \centering
    \includegraphics[width=0.8\linewidth]{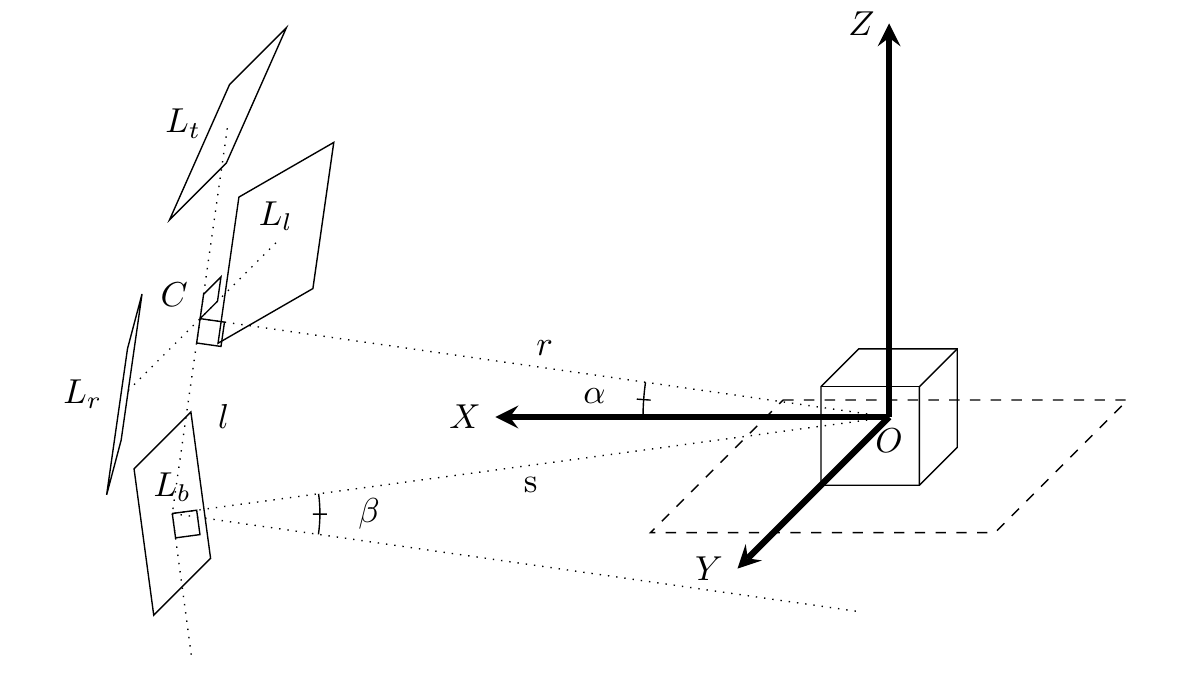}
    \caption{
        An axonometric view of the camera and the surrounding lights. The object, represented as a cube, is placed at the center of the world ($O$). The camera ($C$) faces the object at a distance $r$ making an angle $\alpha$ with the $X$ axis. Four square-shaped lights ($L_t$, $L_r$, $L_b$ and $L_l$) are placed on a circle of radius $l$, centered at $C$, at intervals of 90\textdegree. Every light is tilted of an angle $\beta$ in order to face the center of the world. The value of $\beta$ depends on the distances $r$ and $l$.
    }
    \label{fig:cameralights}
\end{figure}

\begin{figure}
    \includegraphics[width=\linewidth]{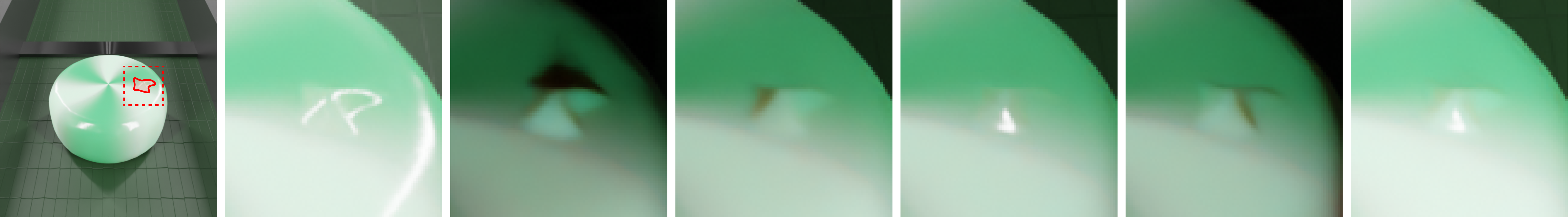}
    \caption{
        An anomalous sample from \emph{Peppermint Candy} category with a bump on its surface. The first figure shows the full picture with the anomalous area highlighted in red. The following figures all show the same crop around the bump, but each one with a different lighting condition.
    }
    \label{fig:anomaliesfromlight}
\end{figure}

We designed the Eyecandies dataset to be a full-fledged, multi-purpose source of data, fully exploiting the inherently controlled nature of a synthetic environment. First, we created a virtual scene resembling an industrial conveyor belt passing through a light box. Four light sources are placed on the four corners of the box, illuminating the whole scene. The camera is placed inside the light box, facing the conveyor at an angle, surrounded by other four square-shaped light sources, as depicted in Fig.~\ref{fig:cameralights}). We will hereinafter use the terms ``box lights'' and ``camera lights'' meaning, respectively, the spotlights at the light box corners, and the light sources surrounding the camera. 
Among the many different conceivable lighting patterns, we chose to provide \emph{six} shots for each sample object:
\begin{enumerate}
    \item one with box lights only.
    \item four with only one camera light (one per light).
    \item one with all camera lights at the same time.
\end{enumerate}
To the best of our knowledge, no other existing dataset for anomaly detection include multiple light patterns for each object,
hence paving the way for novel and exciting approaches to solve the anomaly detection task. Indeed, as shown in Fig.~\ref{fig:anomaliesfromlight}, strong directional shadows enable the detection of surface irregularities otherwise hard to see: while the presence of a defect can appear unclear under some specific lighting conditions, when comparing multiple shots of the same object with different lighting, the detection becomes much easier. 

\begin{figure}
    \centering
    \begin{subfigure}[b]{0.22\textwidth}
        \centering
        \includegraphics[width=\textwidth]{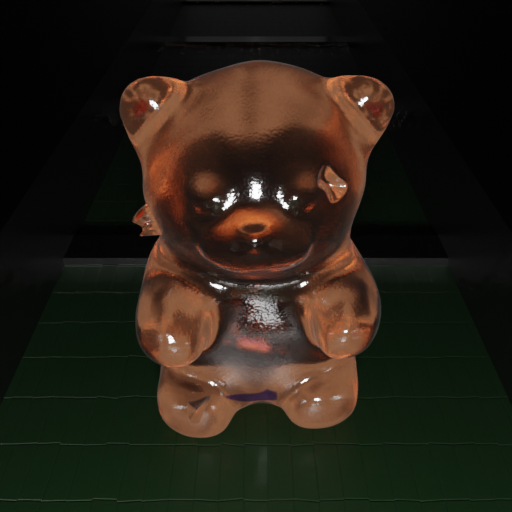}
        \caption{}
        \label{fig:a}
    \end{subfigure}
    \hfill
    \begin{subfigure}[b]{0.22\textwidth}
        \centering
        \includegraphics[width=\textwidth]{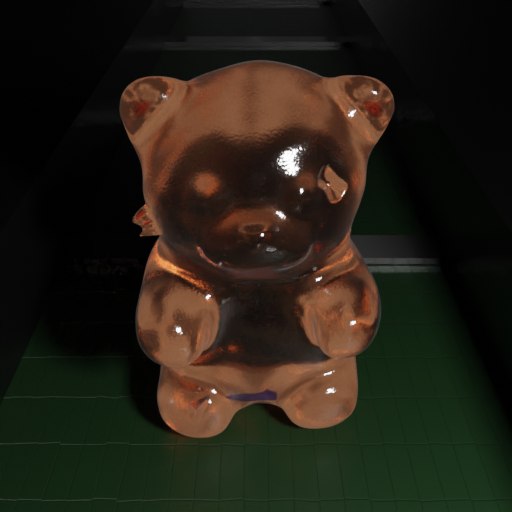}
        \caption{}
        \label{fig:b}
    \end{subfigure}
    \hfill
    \begin{subfigure}[b]{0.22\textwidth}
        \centering
        \includegraphics[width=\textwidth]{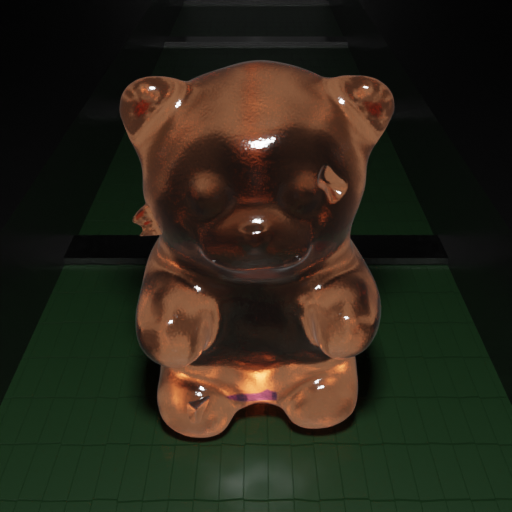}
        \caption{}
        \label{fig:c}
    \end{subfigure}
    \hfill
    \begin{subfigure}[b]{0.22\textwidth}
        \centering
        \includegraphics[width=\textwidth]{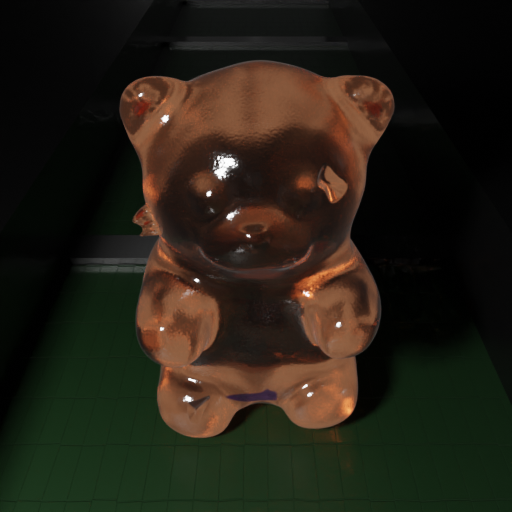}
        \caption{}
        \label{fig:d}
    \end{subfigure}
    \begin{subfigure}[b]{0.22\textwidth}
        \centering
        \includegraphics[width=\textwidth]{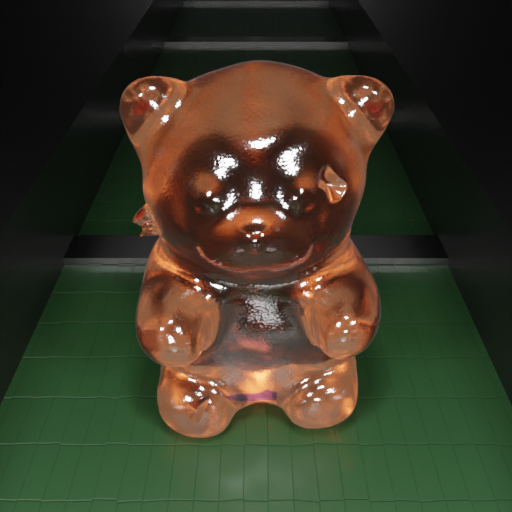}
        \caption{}
        \label{fig:e}
    \end{subfigure}
    \hfill
    \begin{subfigure}[b]{0.22\textwidth}
        \centering
        \includegraphics[width=\textwidth]{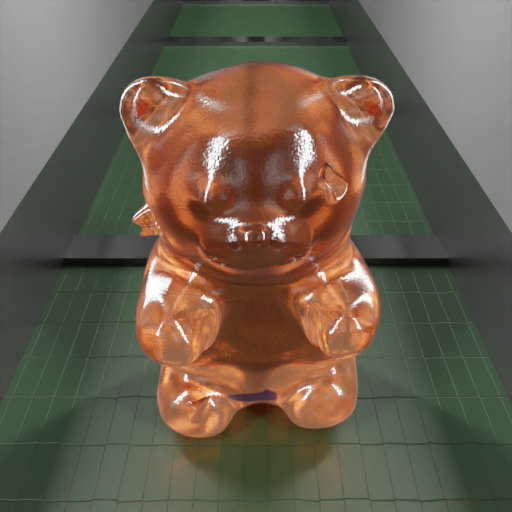}
        \caption{}
        \label{fig:f}
    \end{subfigure}
    \hfill
    \begin{subfigure}[b]{0.22\textwidth}
        \centering
        \includegraphics[width=\textwidth]{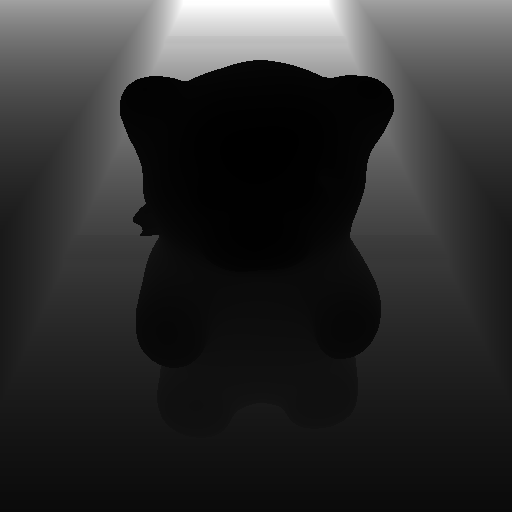}
        \caption{}
        \label{fig:g}
    \end{subfigure}
    \hfill
    \begin{subfigure}[b]{0.22\textwidth}
        \centering
        \includegraphics[width=\textwidth]{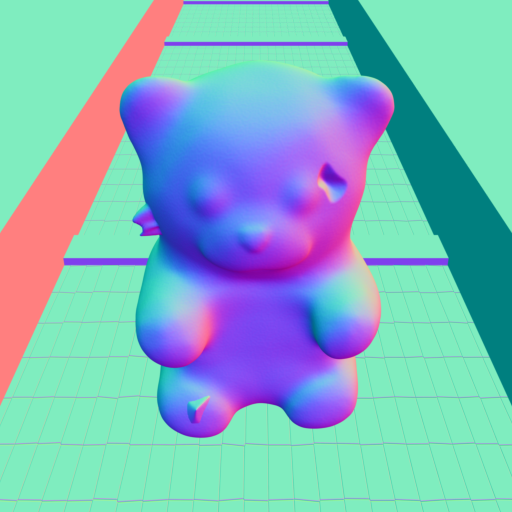}
        \caption{}
        \label{fig:h}
    \end{subfigure}
    \caption{
        Image data of a \emph{Gummy Bear} sample from the test set. Renderings with a single light source are shown in Fig.~\subref{fig:a}--\subref{fig:d}. In Fig.~\subref{fig:e} all the camera lights are active. In Fig.~\subref{fig:f}, the light comes instead from four spotlights at the corners of the surrounding lightbox, and the camera lights are switched off. Fig.~\subref{fig:g} and \subref{fig:h} show the rendered depth and normal maps.
    }
    \label{fig:depthandnormals}
\end{figure}

Alongside RGB color images, depth and normal maps are rendered for each scene. Both are computed by ray-casting the mesh from the same camera point-of-view, i.e., the \emph{depth} is the Z coordinate in camera reference frame, likewise normal directions are expressed as unit vectors in the camera reference frame as well (see an example in Fig.~\ref{fig:depthandnormals}).
The benefit of considering these additional data sources will be discussed in Sec.~\ref{sec:experiments}, where we will show that simply concatenating color, depth and surface normals can boost the performance of a naive autoencoder.
Interestingly, combining depth, normal and RGB images rendered in multiple lighting scenarios allows for addressing many more tasks than just anomaly detection and localization, such as photometric normal estimation and stereo reconstruction, depth from mono and scene relighting.

Since we do not aim at simulating the acquisition from any real devices, no noise is intentionally added to the data. Also, we do not claim that training models on our synthetic dataset may somehow help solving real-world applications. Instead, we provide a clean, though challenging, benchmark to fairly compare existing and future proposals.

\subsection{Synthetic Defects\label{sec:defects}}

Real-world defects come in various shapes and appearances, often tied to particular object features and to the production process. However, we identified common properties and decided to focus on three general groups of anomalies that can occur on many different types of objects:
\begin{enumerate}
    \item color alterations, such as stains and burns;
    \item shape deformations, i.e., bumps or dents;
    \item scratches and other small surface imperfections.
\end{enumerate}
All these groups can be viewed as local anomalies applied on different input data, i.e., color alterations change the RGB image, shape deformations modify the 3D geometry and surface imperfections only alter the normal directions. We chose to include defects that modify the surface normals without affecting the 3D geometry to represent small imperfections hardly captured on a depth map, like, for example, a scratch on a metallic surface. Therefore, we actively change the mesh only in case of surface bumps or dents, while modifications to the normal map have a clear effect only on how the light is reflected or refracted. Finally, for each local defect we provide a corresponding pixel-wise binary mask, rendered directly from the 3D object model, to highlight the area a detector should identify as anomalous.

We purposely left for future investigations the inclusion of two class of anomalies.
Firstly, we avoided class-specific defects, which would have been relevant for only one object class. An example might be altering the number of stripes of the \emph{Marshmallow} or changing the text printed on the \emph{Confetto}. However, this would introduce much more effort in designing how such anomalies should interact with the rendering pipeline, while current defects can be applied in an automated way with no prior information, as described in Sec.~\ref{sec:data-gen}. Secondly, we did not include \emph{logical} anomalies, as described in \cite{MVTECADLOGIC}, because we believe there is no clear consensus on how to evaluate the localization performance of a detector on, e.g., finding missing object regions, and neither on how to annotate such regions in ground-truth anomaly masks.

Unlike many existing dataset, we do not require any human intervention, thus removing a possible source of biases and inconsistencies.

\section{Data Generation\label{sec:data-gen}}

We generated the Eyecandies dataset within the \emph{Blender} framework \cite{blenderman}, a popular 3D modeling software that provides a good level of interoperability with the Python programming language via the \emph{BlenderProc} package \cite{BLENDERPROC}, a useful tool for procedural synthetic data generation.
Every candy class is modelled as a parametric \textit{prototype}, i.e., a rendering pipeline where geometry, textures and materials are defined through a programming language. In this way, object features can be controlled by a set of scalar values, so that each combination of parameters leads to a unique output.

To generate different object instances, we treat all parameters as uniformly distributed random variables, where bounds are chosen to produce a reasonable variance among them, though preserving a realistic aspect for each candy.
Furthermore, to achieve a high degree of photo-realism as well as intra-class variance, key to any procedural object generation pipeline is the use of \emph{noise textures}, useful to produce slight random deformations (as in \emph{Licorice Sandwich}) or fine-grained irregularities in a rough surface. Unlike the parameters described above, these textures are controlled by setting a generic \emph{random seed}, hence letting blender generate them within a geometry modifier or shader. No other noise source is intentionally added, e.g., we do not model any real acquisition device. However, since Cycles, the chosen rendering engine, inevitably introduces random imperfections when computing surface colors, we counteract them by applying NVIDIA Optix\texttrademark{} de-noising \cite{nvidia_optix}.

Anomalous samples differ in that they are given four more textures as parameters, one for each possible defect, i.e., following Sec.~\ref{sec:defects}, \emph{colors}, \emph{bumps}, \emph{dents} and \emph{normals}.
In order to get a realistic appearance, these textures are mapped onto the object mesh through a UV mapping and rendered as part of the object. However, non-trivial constraints must be applied, since simply generating a random blob on a black background would not suffice: it could end up outside of UV map islands, having no effect on the object, or worse, on the edge between two islands, resulting in multiple anomalies with spurious shape. Instead, we export the original UV map of the object from blender, compute a binary mask of the valid areas, then separate all connected components and for each of them compute the maximum inbound rectangle. A random blob is then generated and stitched into one of the inbound rectangles, chosen randomly. This ensures that the anomalies are always entirely visible and never lying on the border of a UV map island.
The quality achieved can be appreciated in Fig.~\ref{fig:allexamples}: modifying the 3D model produces more realistic images than artificially applying the defects on the 2D renderings, while still being able to auto-generate pixel-wise ground-truth masks with no human intervention.

Every object category contains a total of 1500 samples, split between training, validation and test sets consisting of, respectively, 1000, 100 and 400 samples. Training and validation sets provide good examples only, whereas half of the test samples are defective candies. Also, these 200 anomalous samples contain a balanced mixture of the four anomaly types: 40 examples for each one totalling to 160 examples, the remaining 40 containing all possible anomalies together.

\section{Experiments\label{sec:experiments}}

First, we evaluated existing methods for AD on Eyecandies and compared with the results obtained on MVTec AD \cite{MVTECAD}. In Tab.~\ref{table:imageAUROC-others} the area under ROC curve (AUROC) is reported for Ganomaly (G) \cite{akcay2018ganomaly}, Deep Feature Kernel Density Estimation (DFKDE) \cite{dfkde}, Probabilistic Modeling of Deep Features (DFM) \cite{dfm}, Student-Teacher Feature Pyramid Matching (STFPM) \cite{STUDENT-TEACHER} and PaDiM \cite{PADIM}, all run within the Anomalib framework \cite{anomalib}. We notice a significant correlation between the performance on the \emph{real} dataset MVTec AD and Eyecandies, thus suggesting that our proposal, though synthetic, is a valid approach to evaluate AD methods. Furthermore, all methods, with the exception of Ganomaly \cite{akcay2018ganomaly}, show a large performance drop when trained and tested on Eyecandies, proving the increased complexity of the task w.r.t.\ popular AD datasets, such as MVTec AD.

\setlength{\tabcolsep}{4pt}
\begin{table}
\begin{center}
\begin{tabular}{m{0.2\textwidth} m{0.054\textwidth} m{0.054\textwidth} m{0.054\textwidth} m{0.054\textwidth} m{0.054\textwidth} m{0.054\textwidth} m{0.054\textwidth} m{0.054\textwidth} m{0.054\textwidth} m{0.054\textwidth}}
\hline\noalign{\smallskip}
\multirow{2}{*}{Category} & \multirow{2}{*}{G\cite{akcay2018ganomaly}} & \multicolumn{2}{c}{DFKDE\cite{dfkde}} & \multicolumn{2}{c}{DFM\cite{dfm}} & \multicolumn{2}{c}{STFPM\cite{STUDENT-TEACHER}} & \multicolumn{2}{c}{PaDiM\cite{PADIM}} & Ours
\\
 &  & r18 & wr50 & r18 & wr50 & r18 & wr50 & r18 & wr50 & RGB
\\
\noalign{\smallskip}
\hline
\noalign{\smallskip}
Candy Cane & 0.485 & 0.537 & 0.539 & 0.529 & 0.532 & 0.527 & 0.551 & 0.537 & 0.531 & 0.527
\\
Chocolate C. & 0.512 & 0.589 & 0.577 & 0.759 & 0.776 & 0.628 & 0.654 & 0.765 & 0.816 & 0.848
\\
Chocolate P. & 0.532 & 0.517 & 0.482 & 0.587 & 0.624 & 0.766 & 0.576 & 0.754 & 0.821 & 0.772
\\
Confetto & 0.504 & 0.490 & 0.548 & 0.649 & 0.675 & 0.666 & 0.784 & 0.794 & 0.856 & 0.734
\\
Gummy Bear & 0.558 & 0.591 & 0.541 & 0.655 & 0.681 & 0.728 & 0.737 & 0.798 & 0.826 & 0.590
\\
Hazelnut T. & 0.486 & 0.490 & 0.492 & 0.611 & 0.596 & 0.727 & 0.790 & 0.645 & 0.727 & 0.508
\\
Licorice S. & 0.467 & 0.532 & 0.524 & 0.692 & 0.685 & 0.738 & 0.778 & 0.752 & 0.784 & 0.693
\\
Lollipop & 0.511 & 0.536 & 0.602 & 0.599 & 0.618 & 0.572 & 0.620 & 0.621 & 0.665 & 0.760
\\
Marshmallow & 0.481 & 0.646 & 0.658 & 0.942 & 0.964 & 0.893 & 0.840 & 0.978 & 0.987 & 0.851
\\
Peppermint C. & 0.528 & 0.518 & 0.591 & 0.736 & 0.770 & 0.631 & 0.749 & 0.894 & 0.924 & 0.730
\\
\noalign{\smallskip}
\hline
\noalign{\smallskip}
Avg. Eyecandies & 0.507 & 0.545 & 0.555 & 0.676 & 0.692 & 0.688 & 0.708 & 0.754 & 0.794 & 0.701
\\
Avg. MVTecAD & 0.421 & 0.762 & 0.774 & 0.894 & 0.891 & 0.893 & 0.876 & 0.891 & 0.950
\\
\hline
\end{tabular}
\end{center}
\caption{
Image AUROC of existing AD methods on Eyecandies Dataset, considering only RGB images. We compared feature-based methods with both Resnet18 (r18) and Wide-Resnet50 (wr50) pre-trained backbones.
}
\label{table:imageAUROC-others}
\end{table}
\setlength{\tabcolsep}{1.4pt}

As for understanding the contribution of the 3D data to the AD task, we trained a different deep convolutional autoencoder on each of the 10 categories of the Eyecandies dataset. The model has been trained to reconstruct defect-free examples and evaluated on the test set containing a mixture of good and bad examples. Since the network sees only defect-free data during training, we expect anomalous areas to give large reconstruction errors. Thus, beside image-wise metrics, a per-pixel anomaly score can be computed as the L1 distance between the input image and its reconstruction, averaging over the image channels. Similarly, the image-wise anomaly score is computed as the maximum of the per-pixel score.

\begin{figure}
    \includegraphics[width=\linewidth]{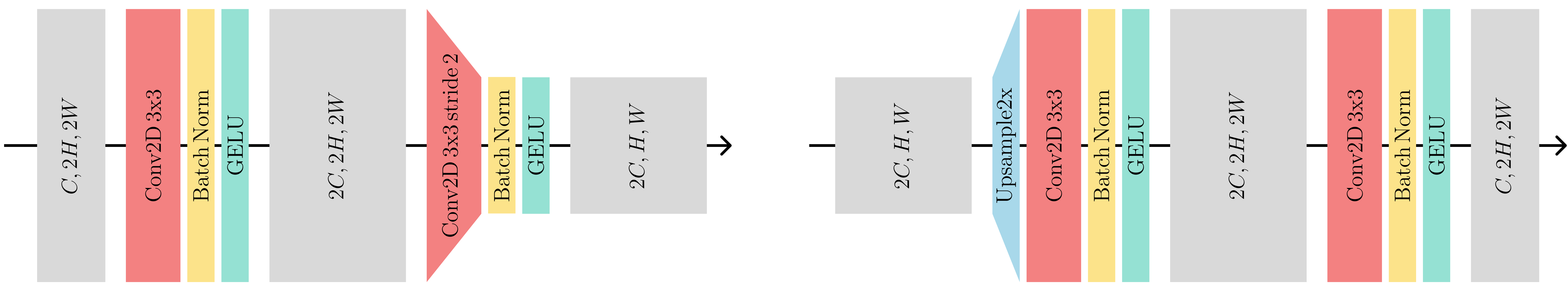}
    \caption{\label{fig:blocks}
        The encoder (left) and decoder (right) blocks inner structure.
    }
\end{figure}

The model consists of two symmetrical parts, i.e., the encoder and the decoder, connected by a linear fully-connected bottleneck layer. Every encoder block increases the number of filters by a factor of 2 with respect to the previous one and halves the spatial resolution by means of strided convolutions. Each decoder block, on the other hand, halves the number of filters while doubling the spatial resolution. Both the encoder and the decoder comprise 4 blocks and the initial number of filters is set to 32, hence, starting from an input of size $3\times256\times256$, the bottleneck layer is fed with a $256\times16\times16$ tensor which is projected onto a latent space with 256 dimensions. Then, the decoder expands this feature vector back to $256\times16\times16$ and up to the same initial dimensions. The inner block layers are detailed in Fig.~\ref{fig:blocks}.

For every object category, we trained the autoencoder with different input combinations: RGB, RGB + Depth (RGBD), RGB + Depth + Normals (RGBDN), all downscaled to a fixed resolution of $256\times256$ pixels.
To this end, being the color image as well as the depth and the normal maps of equivalent resolution, we simply concatenated them along the channel dimension, changing the number of input and output channels of the autoencoder accordingly. Therefore, the total number of channels is 3 for the RGB case, 4 when adding the depth and 7 when using all the available inputs, i.e., color, depth and normals. When using the depth maps, the values are re-scaled between 0 and 1 with a per-image min-max normalization. As for data augmentation, we used the following random transforms:
\begin{itemize} 
    \item random shift between -5\% and 5\% with probability 0.9;
    \item random scale between 0\% and 5\% with probability 0.9;
    \item random rotation between -2\textdegree and 2\textdegree with probability 0.9;
    \item HSV color jittering of 5\textdegree with probability 0.9 (RGB only).
\end{itemize} 

Following \cite{AutoencoderSSIM}, we defined the loss function as follows:

\begin{equation}
  L\left( I, \hat I\right) = L_{L1}\left( I, \hat I \right) + L_{SSIM}\left( I, \hat I \right)
  \label{eq:loss}
\end{equation}

where $I$ is the autoencoder input, $\hat I$ its reconstruction, $L_{L1}\left( I, \hat I \right) =\left\| I- \hat I \right\|_1$ is the reconstruction error and $L_{SSIM}$ the Multi-Scale Structural Similarity as defined in \cite{AutoencoderSSIM}. The SSIM window size is set to 7 pixels. When RGB, depth and normals maps are concatenated, the loss is computed as the sum of all the individual loss components:

\begin{equation}
    L_{RGBDN}\left( I, \hat I\right) = L\left( RGB, \hat{RGB}\right) + L\left( D, \hat D\right) + L\left( N, \hat N\right)
    \label{eq:loss-rgbdn}
\end{equation}

Where $RGB$, $D$ and $N$ are, respectively, the color image, the depth map and the normals map, each reconstructed as $\hat{RGB}$, $\hat D$ and $\hat N$. All models are trained for 5000 epochs with Adam \cite{kingma2014adam} optimizer with learning rate 0.001, $\beta_1$ set to 0.9 and $\beta_2$ to 0.999. The mini-batch size is set to 32, enforced by dropping the last batch of every epoch if the required batch size is not met.

\begin{figure}
    \includegraphics[width=\linewidth]{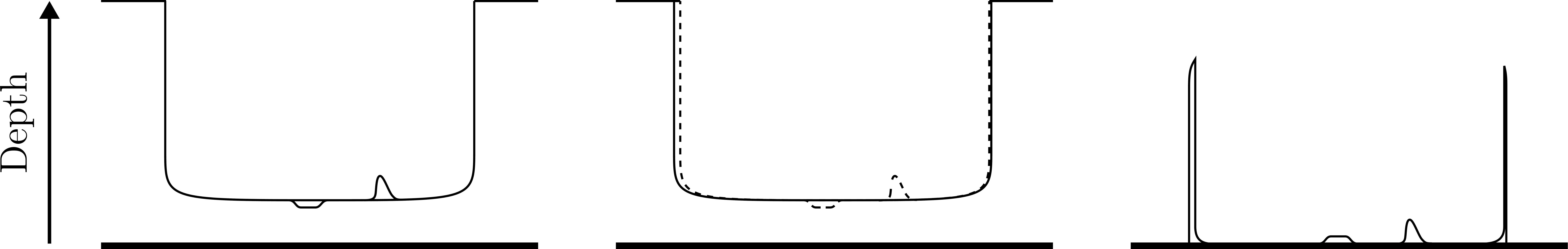}
    \caption{\label{fig:deptherror}
        On the left, the height map of a row from the depth image of an anomalous sample, containing a bump and a dent on its surface. On the center, its reconstruction with no surface anomalies, but with a slightly misaligned contour (the dashed line represents the original depth). On the right, the absolute difference of the two. The reconstruction error in the anomalous areas is negligible with respect to the error on the object contour.
    }
\end{figure}

The results, summarized in Tab.~\ref{table:imageAUROC} and Tab.~\ref{table:pixelAUROC} as, respectively, image and pixel AUROC scores, suggest that a naive autoencoder trained on RGB data alone fails to effectively separate \emph{good} and \emph{bad} samples on most object categories. The worst performance is reached on \emph{Candy Cane} and \emph{Hazelnut Truffle}, where the results are comparable to a random classifier. In the former case, though the reconstruction quality is good, the anomalies may be too small to be effectively detected. In the latter, the detection possibly fails because of the low reconstruction quality, due to the roughness of the surface and known issues of convolutional autoencoders with high frequency spatial features \cite{AutoencoderSSIM}.
Conversely, \emph{Gummy Bear} exhibits a smooth and glossy surface, where edges are surrounded by bright glares, making the reconstruction from color images hardly conceivable.
In fact, we achieve acceptable results only on objects with regular and opaque surfaces, thus easier to reconstruct, such as \emph{Chocolate Cookie} and \emph{Marshmallow}, where, moreover, anomalies appear on a fairly large scale.

Adding the depth map to the reconstruction task has little impact on the network performance.
To understand the diverse reasons behind this, first consider that only two types of anomalies, namely bumps and dents, have an effect on the object depth, since color and normal alterations do not affect the geometry of the object.
Furthermore, as depicted in Fig.~\ref{fig:deptherror}, the average height of a bump or a dent is very small if compared to the size of the object, so that the reconstruction error of the anomalous regions is often negligible with respect to the error found on the silhouette of the object, where a naive autoencoder often produces slightly misaligned shapes. Also, extending the input and the output with depth information, the task becomes more complex and the reconstruction quality of the color image worsens when compared to the RGB-only scenario. This, eventually, leads to no improvement or even a drop in performance for most of the objects, being the color texture usually more informative.

Introducing the normals into the reconstruction task drastically improves the performance in almost every object category. Although normal maps do not provide any meaningful information to detect color alteration anomalies, they prove to be crucial to detect bumps, dents and surface normal alterations. Moreover, unlike the RGB image, normal maps do not suffer from glossy materials, reflections and sharp edged in the object texture, hence easing the reconstruction task.

We made an attempt to mitigate the problem with the low visibility of anomalies on the error maps obtained from depth reconstruction (the one described in Fig.~\ref{fig:deptherror}), by clamping the depth maps between fixed lower and upper bounds. This helps reducing the contrast - and thus the reconstruction error - between pixels on the object silhouette and the background. Bounds were chosen so that the final depth range would be as small as possible, with the constraint to never affect pixels belonging to the object: the lower bound is always smaller than the closest point, the upper bound is always greater than the farthest visible point.

We repeated the RGB-D and RGB-D-N experiments described before, under the exact same settings, but clamping the depth maps. Results are summarized in the columns ``RGB-cD'' and ``RGB-cD-N'' of Tab.~\ref{table:imageAUROC} and Tab.~\ref{table:pixelAUROC}. As for RGB-D and RGB-cD, we can observe that the depth clamping results in slightly better performances across most object categories, both for image and pixel AUROC. When using also normal maps, the benefits of having a clamped depth range seem to have less relevance: only 6 categories out of 10 see an improvement in image AUROC, but pixel metrics improve only in 2 cases.

\setlength{\tabcolsep}{4pt}
\begin{table}
\begin{center}
\begin{tabular}{llllll}
\hline\noalign{\smallskip}
Category & RGB & RGB-D & RGB-D-N & RGB-cD & RGB-cD-N \\
\noalign{\smallskip}
\hline
\noalign{\smallskip}
Candy Cane & 0.527 & 0.529 & 0.587 & 0.537 & \bf 0.596
\\
Chocolate Cookie & 0.848 & \bf 0.861 & 0.846 & 0.847 & 0.843
\\
Chocolate Praline & 0.772 & 0.739 & 0.807 & 0.748 & \bf 0.819
\\
Confetto & 0.734 & 0.752 & 0.833 & 0.779 & \bf 0.846
\\
Gummy Bear & 0.590 & 0.594 & \bf 0.833 & 0.635 & \bf 0.833
\\
Hazelnut Truffle & 0.508 & 0.498 & 0.543 & 0.511 & \bf 0.550 
\\
Licorice Sandwich & 0.693 & 0.679 & 0.744 & 0.691 & \bf 0.750
\\
Lollipop & 0.760 & 0.651 & \bf 0.870 & 0.699 & 0.846
\\
Marshmallow & 0.851 & 0.838 & \bf 0.946 & 0.871 & 0.940
\\
Peppermint Candy & 0.730 & 0.750 & 0.835 & 0.740 & \bf 0.848
\\
\end{tabular}
\end{center}
\caption{
Autoencoder Image AUROC on the Eyecandies Dataset.
}
\label{table:imageAUROC}
\end{table}
\setlength{\tabcolsep}{1.4pt}

\setlength{\tabcolsep}{4pt}
\begin{table}
\begin{center}
\begin{tabular}{llllll}
\hline\noalign{\smallskip}
Category & RGB & RGB-D & RGB-D-N & RGB-cD & RGB-cD-N \\
\noalign{\smallskip}
\hline
\noalign{\smallskip}
Candy Cane & 0.972 & 0.973 & \bf 0.982 & 0.975 & 0.980
\\
Chocolate Cookie & 0.933 & 0.927 & \bf 0.979 & 0.939 & \bf 0.979
\\
Chocolate Praline & 0.960 & 0.958 & 0.981 & 0.954 & \bf 0.982
\\
Confetto & 0.945 & 0.945 & \bf 0.979 & 0.957 & 0.978
\\
Gummy Bear & 0.929 & 0.929 & \bf 0.951 & 0.933 & \bf 0.951
\\
Hazelnut Truffle & 0.815 & 0.806 & 0.850 & 0.822 & \bf 0.853 
\\
Licorice Sandwich & 0.855 & 0.827 & \bf 0.972 & 0.897 & 0.971
\\
Lollipop & 0.977 & 0.977 & \bf 0.981 & 0.978 & 0.978
\\
Marshmallow & 0.931 & 0.931 & \bf 0.986 & 0.940 & 0.985
\\
Peppermint Candy & 0.928 & 0.928 & \bf 0.967 & 0.940 & \bf 0.967
\\

\end{tabular}
\end{center}
\caption{
Autoencoder Pixel AUROC on the Eyecandies Dataset.
}
\label{table:pixelAUROC}
\end{table}
\setlength{\tabcolsep}{1.4pt}


\section{Conclusions And Future Works\label{sec:conclusions}}

We presented Eycandies, a novel synthetic dataset for anomaly detection and localization. Unlike existing datasets, for each unique object instance we provide RGB color images as well as depth and normals maps, captured under multiple light conditions. We have shown how unique candies are generated from a parametric reference model, with photo-realistic appearance and large intra-class variance. Likewise, random anomalies are carefully applied on color, depth and normals, then reprojected to 2D to get pixel-precise ground-truth data, avoiding any human intervention. Our experiments suggest that combining color and 3D data may open new possibilities to tackle the anomaly detection task and our dataset might be crucial to validate such new results.

As for future works, we see four main subjects to investigate. Firstly, we should expand the dataset by moving the camera around the target object, hence generating a full view. Secondly, we might add logical anomalies, such as  missing parts, together with a sensible ground truth and a clear evaluation procedure. Thirdly, we could generate object-specific defects, such as wrong stripes on the \emph{Candy Cane}, though the challenging part would be to leave the whole pipeline completely automated. Finally, we might model the noise, artifacts and deformations introduced by acquisition devices.

%
%
%
\bibliographystyle{splncs04}
\bibliography{egbib}

\end{document}